\documentclass{Interspeech}

\usepackage{comment}
\usepackage{subcaption}
\usepackage{graphicx}
\usepackage[utf8]{inputenc}
\usepackage[T1]{fontenc}
\usepackage{multirow}

\def\F0{$F_0$\xspace}

\interspeechcameraready 

\title{RepeaTTS: Towards Feature Discovery through Repeated Fine-Tuning} 

\author[affiliation={1}]{Atli}{Sigurgeirsson}
\author[affiliation={1}]{Simon}{King}

\affiliation{The Centre for Speech Technology Research}{University of Edinburgh}{UK}

\email{s2063518@ed.ac.uk, simon.king@ed.ac.uk}

\begin{document}

\maketitle

\begin{abstract}
    A \textit{Prompt-based} Text-To-Speech model allows a user to control different aspects of speech, such as speaking rate and perceived gender, through natural language instruction. Although user-friendly, such approaches are on one hand constrained: control is limited to acoustic features exposed to the model during training, and too flexible on the other: the same inputs yields uncontrollable variation that are reflected in the corpus statistics. 

    We investigate a novel fine-tuning regime to address both of these issues at the same time by exploiting the uncontrollable variance of the model. Through principal component analysis of thousands of synthesised samples, we determine latent features that account for the highest proportion of the output variance and incorporate them as new labels for secondary fine-tuning. We evaluate the proposed methods on two models trained on an expressive Icelandic speech corpus, one with emotional disclosure and one without. In the case of the model without emotional disclosure, the method yields both continuous and discrete features that improve overall controllability of the model.  
\end{abstract}

\section{Introduction}

Modern Text-To-Speech (TTS) models can produce very natural-sounding speech by learning the complex mapping between text and acoustic features. Given text, TTS models predict a probable speech rendition under the statistics of the training data. The model can produce countless renditions for the same inputs, which may or may not be perceptually appropriate for the given context. In most cases, TTS models offer no or limited control over how the prosodic rendition is determined: the model produces a probable rendition given just the target text. Models that do offer control over prosodic features \cite[for example]{skerry2018towards, ren2020fastspeech, lyth2024natural} all have one thing in common: the control they offer is limited to the distribution of the features the model has seen during training.

\section{Background}
To some degree, prosody control can be achieved through data-labelling and supervised training given those labels. Training acoustic feature \textit{predictors} jointly with the acoustic model, like the FastSpeech series of models \cite{ren2019fastspeech, ren2020fastspeech}, offers fine-grained control over prosodic correlates. However, controlling prosody in this manner is time-consuming, and controlling learned features independently often leads to degraded quality \cite{mauryahuman}. 

Reference-based models allow for control of certain aspects of the output by conditioning generation on a reference utterance. Reference-conditioning is used for dialogue-based TTS \cite{oplustil2020using}, emotive TTS \cite{van2021exploring}, speaking style modelling \cite{wang2018style, valle2020mellotron} and for \textit{prosody transfer} \cite{skerry2018towards}. Prosody transfer models offer a highly specific control of prosody by conditioning generation on a reference utterance that contains certain desirable prosodic features \cite[e.g]{skerry2018towards}. Using a reference to control prosody might be considered intuitive, as the target rendition can be fully specified in the target medium, namely speech. However, because of how these models are trained, they fail to generalise to cross-text and cross-speaker conditions effectively: the reference must be similar to the target text and speaker \cite{sigurgeirsson2023prosody}, and finding a suitable reference utterance is non-trivial. Reference-based models are also susceptible to feature entanglement, and conditioning on references atypical of the training corpus can lead to unintelligible speech \cite{wang2018style, hsu2018hierarchical}.

\textit{Prompt-based} TTS models allow users to specify the value of a limited set of controllable features through a natural language description \cite[e.g.]{lyth2024natural, lacombe-etal-2024-parler-tts, guo2022prompttts, yang2023instructtts}. These models are trained on a large speech corpus annotated with textual descriptions of features extracted for each training utterance in the corpus. \textit{ParlerTTS} \cite{lacombe-etal-2024-parler-tts} is an implementation of the model proposed in \cite{lyth2024natural}. ParlerTTS is an encoder-decoder architecture that predicts Descript Audio Codec (DAC) tokens \footnote{https://github.com/descriptinc/descript-audio-codec}, given the target text and the description prompt. During training, the target text tokens are prepended to the audio tokens. A frozen text encoder (T5 \cite{t5}) encodes the description prompt, which the model cross-attends to during audio code prediction. The description prompts are automatically generated to include information about the speaker's pitch (monotone or expressive), speaking rate and recording conditions. This is done by automatically annotating a large speech corpus with speaking rate, Signal-to-Noise Ratio (SNR), reverberation estimation, PESQ \cite{rix2001perceptual}, SI-SDR \cite{le2019sdr} and pitch estimation using the PENN library \cite{morrison2023cross}.

Controlling a prompt-based TTS model does not require a complete specification of required acoustic features, nor is a reference utterance needed. However, like other types of controllable TTS models, these models are limited to the features the model has been trained on. Yet, Prompt-based models generate a spectrum of possible renditions for the same inputs, reflecting on variation that the user has no control over.

Hypothetically, the training data for models, such as ParlerTTS, could be enriched with an ever-growing list of acoustic features derived directly from the data to account for this variation. 
However, a complete specification of a large set of different acoustic features would presumably require expert domain knowledge to control the model effectively. Instead, we propose exploiting the model variation to find prosodic features encapsulating several acoustic features. 

The current work explores a novel fine-tuning approach for discovering additional controllable features for such models. The proposed method is based on an extensive analysis of thousands of generated samples to determine the latent features that account for perceptual and controllable variance in the output distribution. We then extract a list of exemplar embeddings used to re-label the training corpus by manually analysing a low-dimensional projection of the generated utterances.

The corpus is then re-enrolled for a secondary fine-tuning of the model, incorporating the new labels. A similar approach could be taken to find clusters in a corpus of real speech. However, utterance-level representations of speech inherently contain information about the linguistic content of the utterance. Therefore, such representations often generalise poorly to diverse text inputs. In the current work, we are interested in finding features that apply to any target text. Our approach circumvents this issue by only considering \textit{fixed inputs}: our analysis set is generated using the same target text, speaker and description prompt, leaving the model free to explore the space of possible renditions for the fixed inputs.

As part of the current work, we release a high-quality version of \textbf{T3-emotion}, a ParlerTTS checkpoint fine-tuned on an Icelandic emotive TTS corpus. Icelandic is used in our work only as a test case. Besides the training data employed, there is nothing language-specific about our experiments or how the model is trained. All our findings are likely to apply to any other target language. Model checkpoints and synthesised samples are available on our demo page\footnote{https://atlisig.github.io/SSW\_supp\_material/index.html}.

\section{Method}
\begin{figure}[ht!]
    \centering
    \begin{subfigure}[b]{0.15\textwidth}
        \centering
        \includegraphics[width=\textwidth]{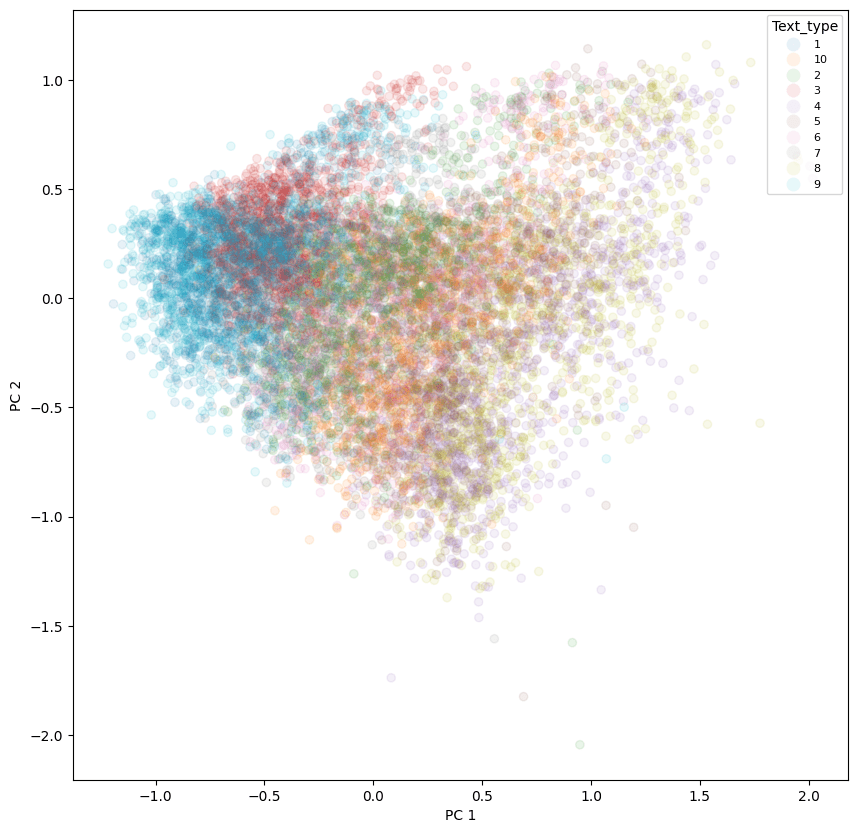}
        \caption*{Layer 1}
    \end{subfigure}
    \hfill
    \begin{subfigure}[b]{0.15\textwidth}
        \centering
        \includegraphics[width=\textwidth]{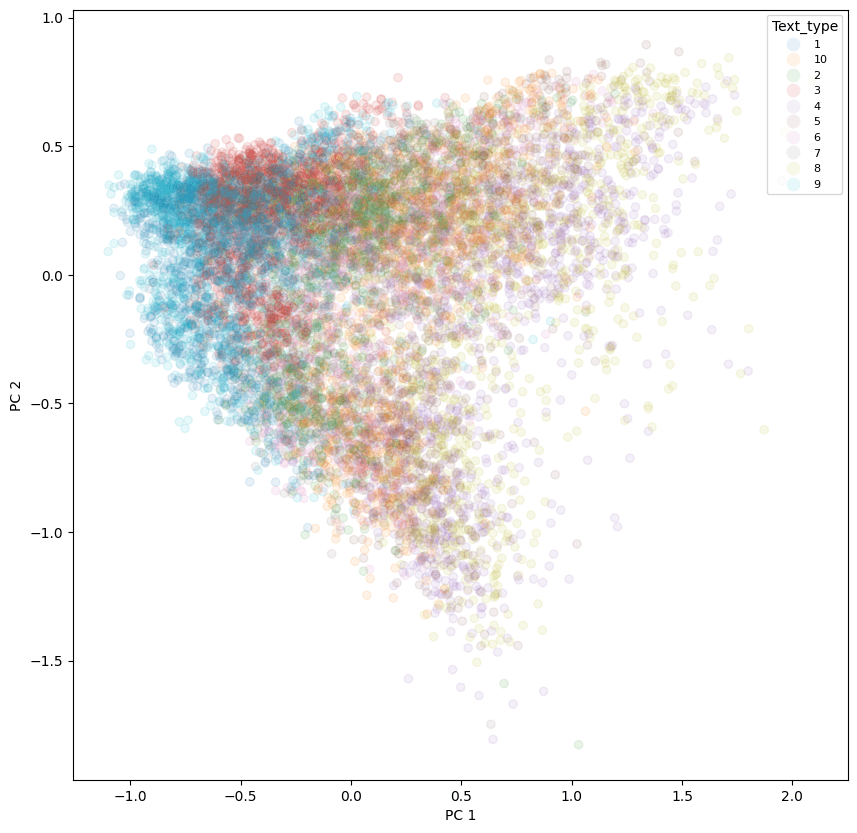}
        \caption*{Layer 2}
    \end{subfigure}
    \hfill
    \begin{subfigure}[b]{0.15\textwidth}
        \centering
        \includegraphics[width=\textwidth]{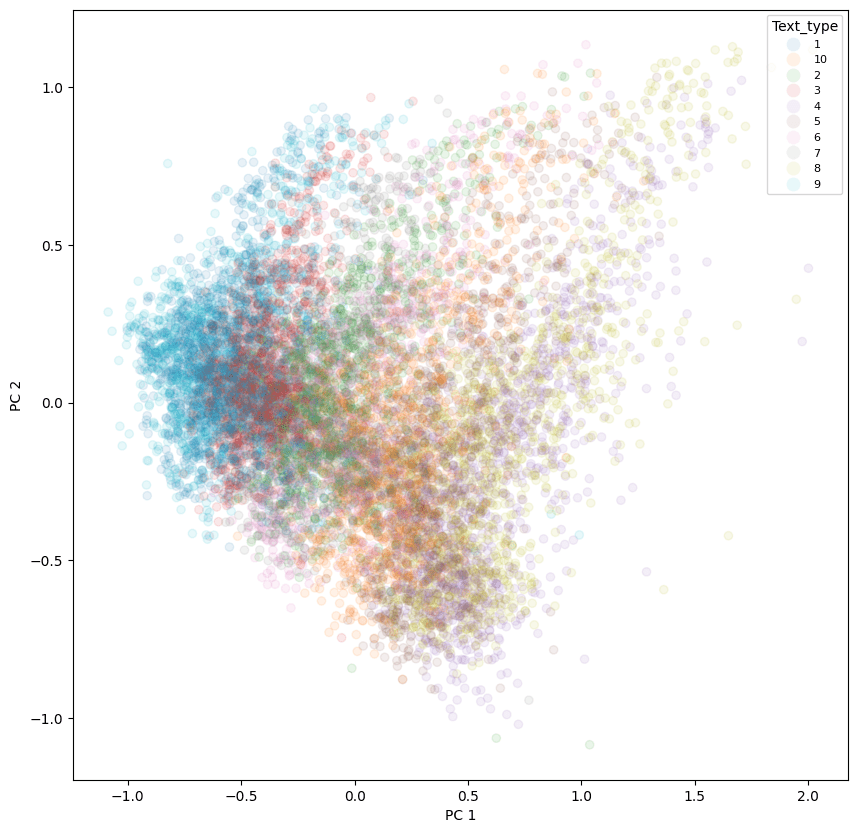}
        \caption*{Layer 3}
    \end{subfigure}
    \hfill
    \begin{subfigure}[b]{0.15\textwidth}
        \centering
        \includegraphics[width=\textwidth]{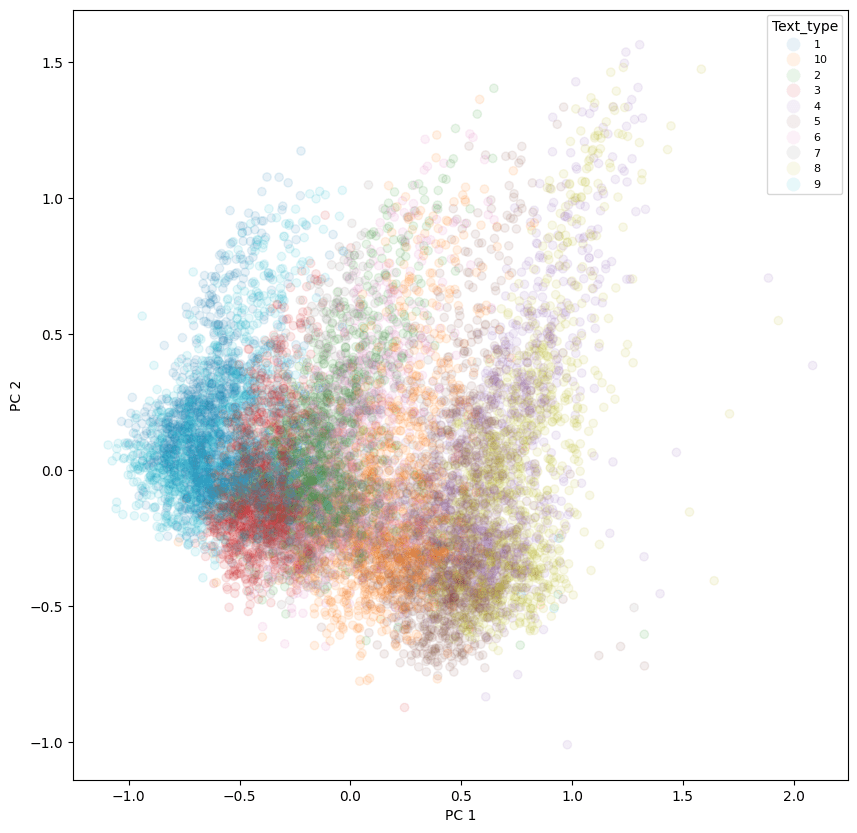}
        \caption*{Layer 4}
    \end{subfigure}
    % start the next row
    \hfill
    \begin{subfigure}[b]{0.15\textwidth}
        \centering
        \includegraphics[width=\textwidth]{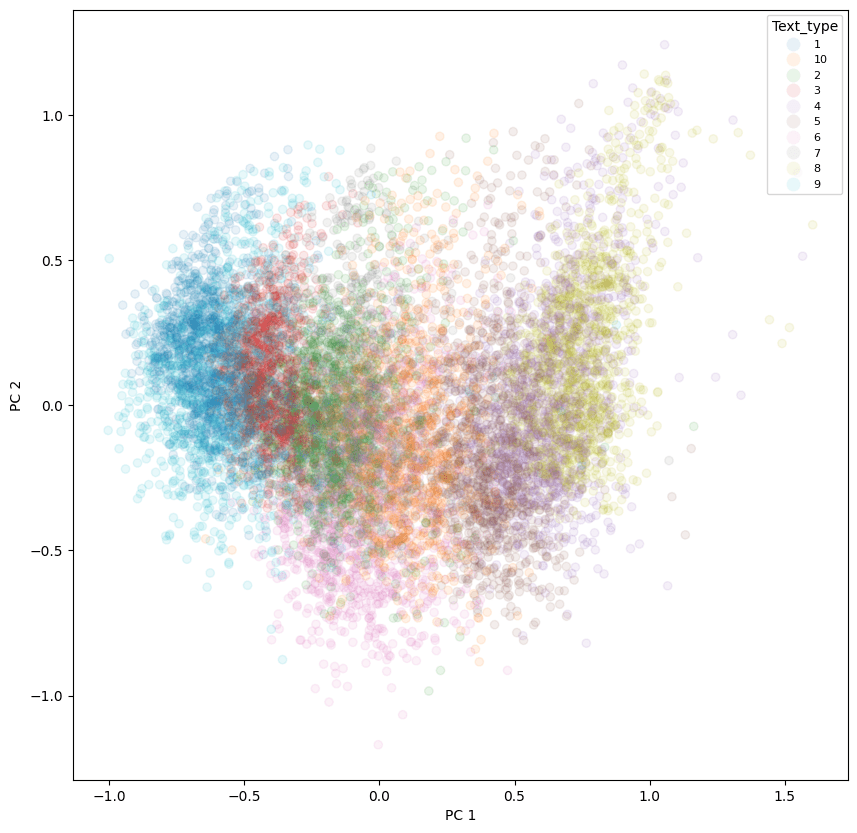}
        \caption*{Layer 5}
    \end{subfigure}
    \hfill
    \begin{subfigure}[b]{0.15\textwidth}
        \centering
        \includegraphics[width=\textwidth]{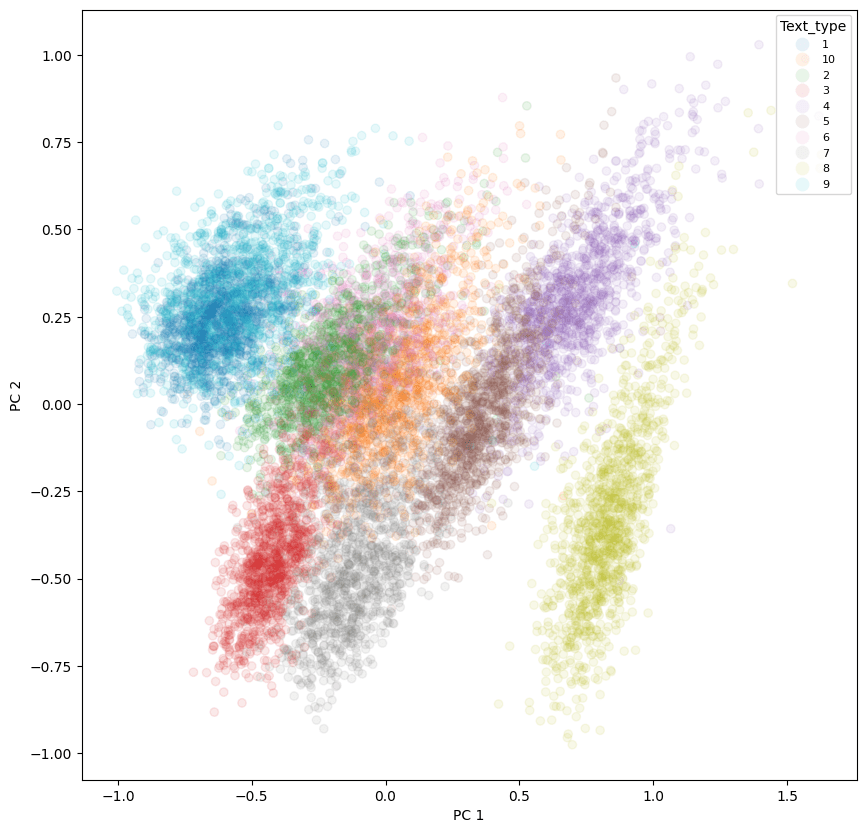}
        \caption*{Layer 6}
    \end{subfigure}
    \hfill
    \begin{subfigure}[b]{0.15\textwidth}
        \centering
        \includegraphics[width=\textwidth]{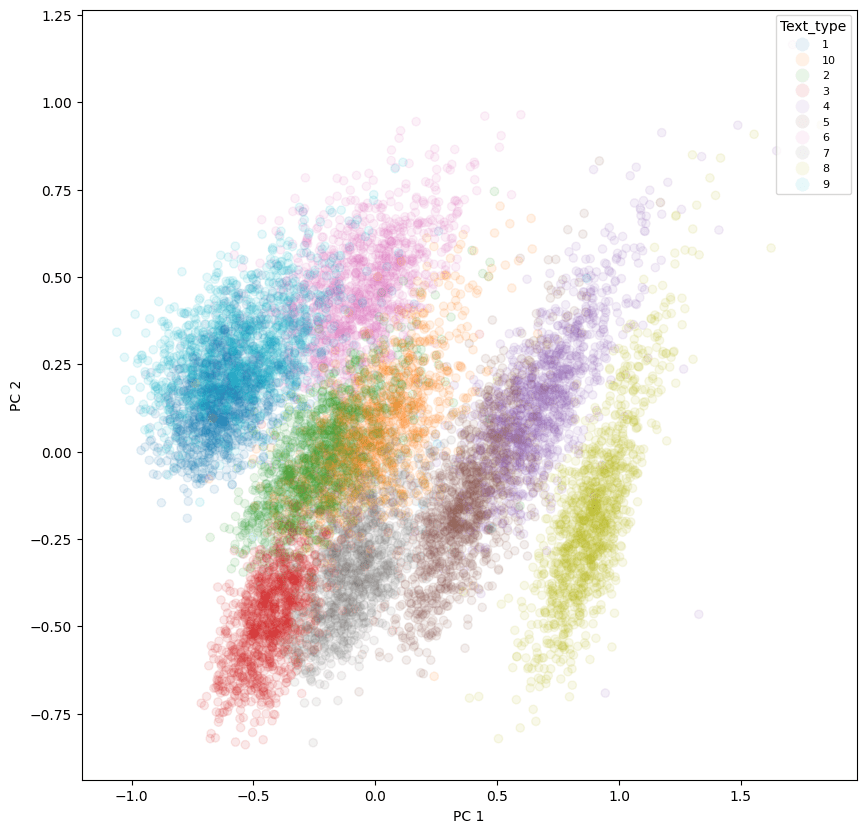}
        \caption*{Layer 7}
    \end{subfigure}
    \hfill
    \begin{subfigure}[b]{0.15\textwidth}
        \centering
        \includegraphics[width=\textwidth]{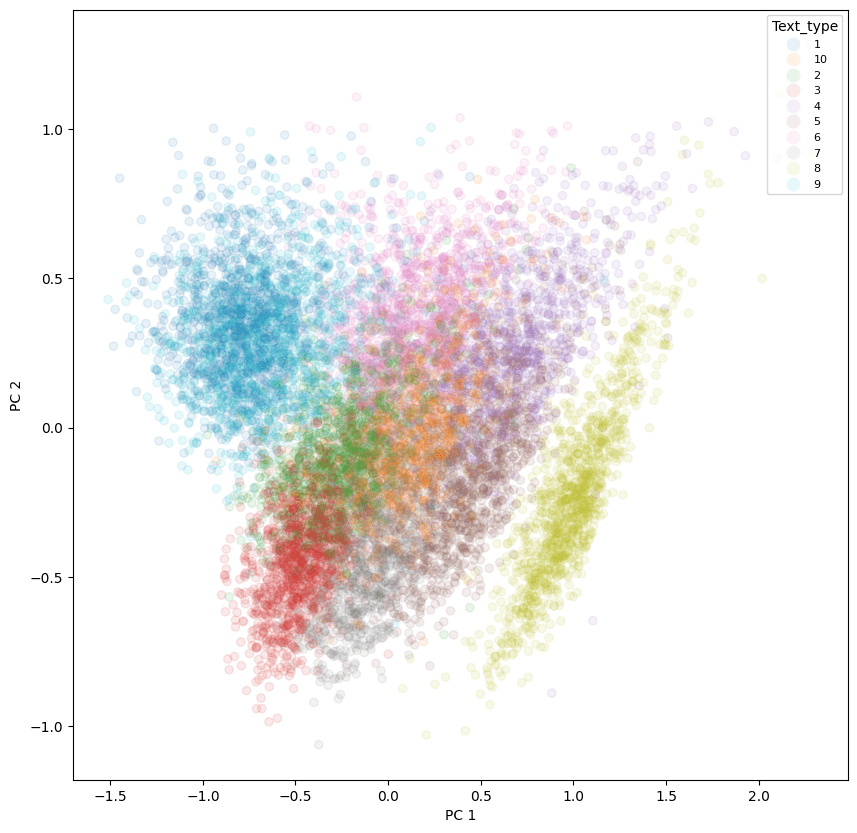}
        \caption*{Layer 8}
    \end{subfigure}
    % start the next row
    \hfill
    \begin{subfigure}[b]{0.15\textwidth}
        \centering
        \includegraphics[width=\textwidth]{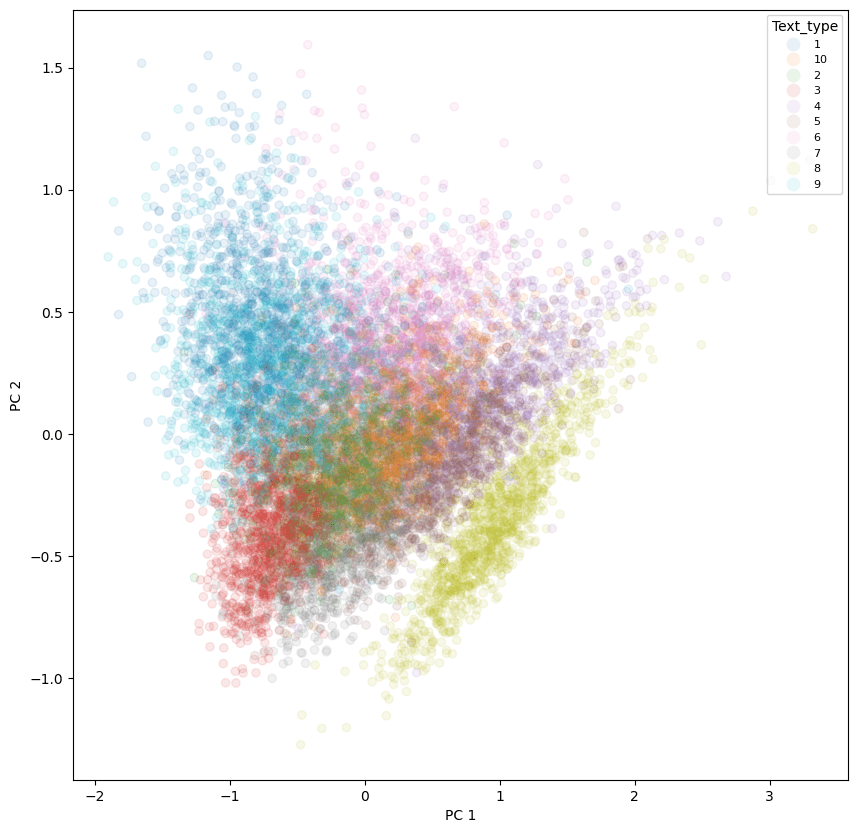}
        \caption*{Layer 9}
    \end{subfigure}
    \hfill
    \begin{subfigure}[b]{0.15\textwidth}
        \centering
        \includegraphics[width=\textwidth]{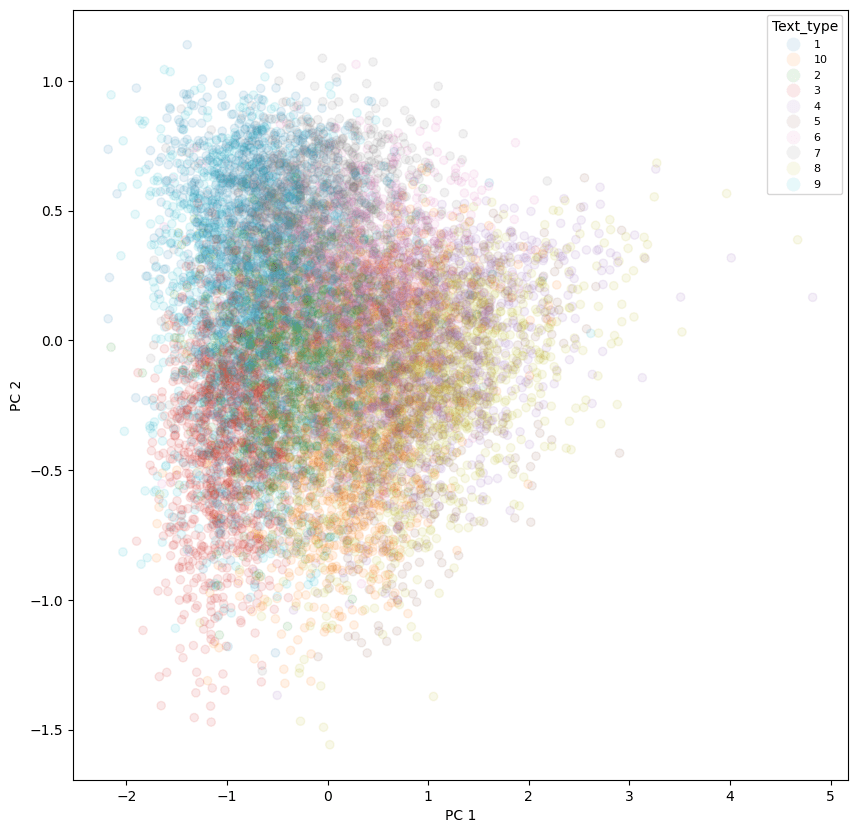}
        \caption*{Layer 10}
    \end{subfigure}
    \hfill
    \begin{subfigure}[b]{0.15\textwidth}
        \centering
        \includegraphics[width=\textwidth]{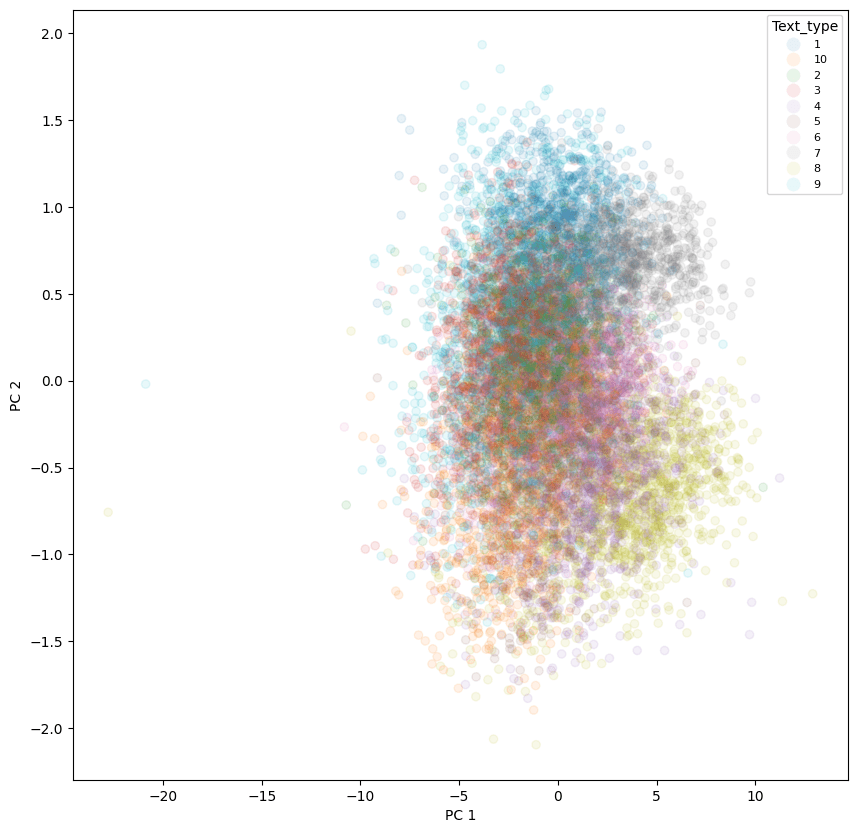}
        \caption*{Layer 11}
    \end{subfigure}
    \hfill
    \begin{subfigure}[b]{0.15\textwidth}
        \centering
        \includegraphics[width=\textwidth]{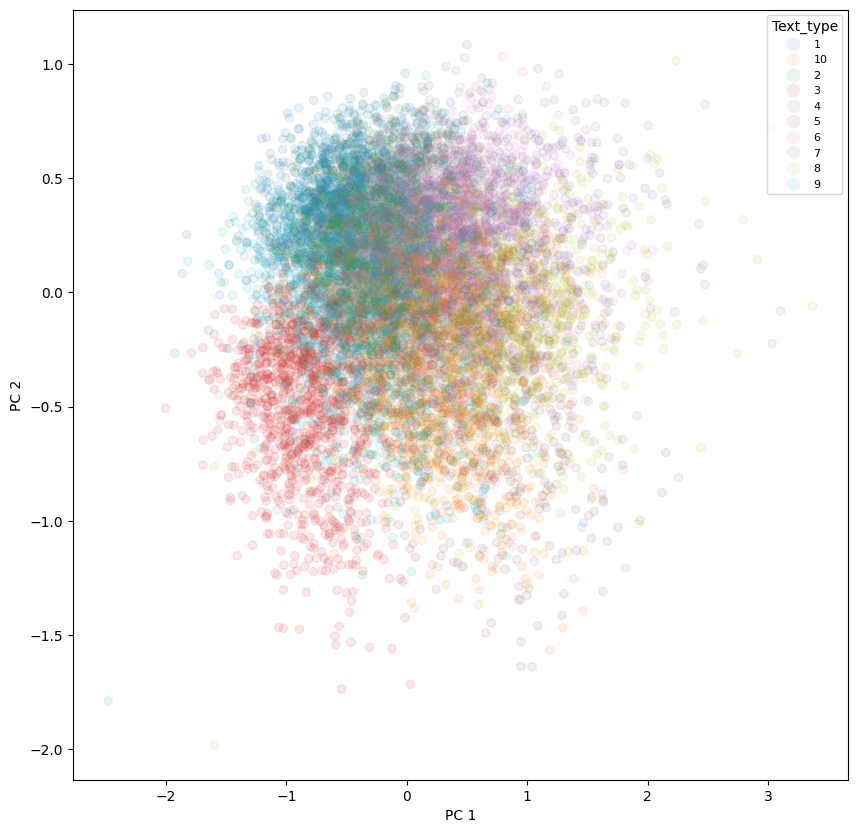}
        \caption*{Layer 12}
    \end{subfigure}
    % add the main caption  
    \caption{PCA of summary embeddings for synthetic speech, colored by different target texts.} \label{fig:diverse_set_everyone_5_pca_text_type}
\end{figure}

This exploratory study can be split into three main phases:
\begin{itemize}
    \item \textbf{Model and representation selection}: We aim to analyse the output distribution of a prompt-based TTS model to identify features that can be enrolled as additional control inputs in a secondary fine-tuning stage. Sections \ref{sec:model}-\ref{sec:representations} discuss which TTS architecture, training data, and utterance representations we use to support the model analysis.
    \item \textbf{Initial model validation}: We start our analysis with two different \textit{baseline} models (\textbf{T3} and \textbf{T3-emotion}) which have been fine-tuned on data in the target language. Our proposed method requires these models to generate diverse renditions for the same input text while adhering to a target speaker label. We objectively evaluate this ability in Section \ref{sec:base-models}.
    \item \textbf{Secondary fine-tuning stages}: We propose to iteratively add new control features --- discovered through the analysis of the model variation --- through sequential fine-tuning of the \textit{baseline} models (\textit{feature enrolment}). We explain the feature exploration method in \ref{sec:feature-exploration} and feature enrollment, which is performed separately for each \textit{baseline} model, is discussed in Sections \ref{sec:t3}-\ref{sec:t3-emotion}.
\end{itemize}

\subsection{Model and Data} \label{sec:model}
We choose ParlerTTS as the model architecture for our experiments. Speaker consistency is a key requirement for our method since we are interested in controlling paralinguistic features independent of the speaker's identity. Checkpoints available in \cite{lacombe-etal-2024-parler-tts} were trained with a list of \textit{known speakers}. The names of those speakers can be included in the description prompt to recall the chosen speaker identity. However, without fine-tuning those checkpoints, we find the speaker identity is frequently inconsistent.  

We, therefore, further fine-tune the model on an additional list of speakers, which are not included in the original training set. We choose to perform our experiments on a high-quality Icelandic emotive speech corpus, \textit{Talromur-3} \cite{talromur3}. We hypothesise that exposing the model to more varied speech than is typical for narration could aid the model in discovering more perceptually salient control features. Talromur-3 is a multi-speaker, emotive speech corpus made specifically for TTS applications. The same list of approximately 400 prompts is spoken according to 5 different emotion labels: happy, sad, angry, surprised and helpful\footnote{can be considered as \textit{child-directed}}. During recording, each utterance is assigned an emotion intensity label on a 5-point Likert scale (e.g. \textit{very low, low, medium, high, very high}). The same list of prompts is also spoken according to a \textit{neutral} emotion label.

To facilitate fine-tuning and experiments on the Icelandic Talromur-3, we employ a multilingual ParlerTTS checkpoint\footnote{https://huggingface.co/parler-tts/parler-tts-mini-multilingual}. Different from the original ParlerTTS model, this checkpoint is trained using LLaMA's tokeniser \cite{touvron2023llama} as a separate prompt tokeniser, which can be extended to other languages such as Icelandic, through byte fallback. Since the model still uses the T5 text encoder, the description prompts remain in English. The checkpoint is trained on eight different European languages, none of which is Icelandic. Yet, initial experiments demonstrated that only limited fine-tuning on Icelandic data is required to synthesise highly natural and intelligible Icelandic.

In our experiments, we initially fine-tune two \textit{baseline} models, \textbf{T3-emotion} and \textbf{T3}. \textbf{T3-emotion} is fine-tuned with textual descriptions of the emotion category and emotional intensity, in addition to a description of the other features that the multilingual checkpoint is trained on. The other model, \textbf{T3}, is fine-tuned on prompts that include no information about the emotion category and must, therefore, learn that mapping from the other features present in the prompt. We use the Dataspeech\footnote{https://github.com/huggingface/dataspeech} repository to generate the description prompts. Dataspeech prompts \texttt{Mistral-7B-Instruct-v0.2} \cite{jiang2023mistral7b} to generate diverse natural language descriptions of the acoustic variation found in the training utterances. We create a new instruction prompt\footnote{available at our demo page.} to include information about the emotional content of the utterances.

\subsection{Utterance Representations} \label{sec:representations}
We extract a diverse set of features for each generated utterance to analyse the model's output distribution. Among those are Wav2Vec2 embeddings, which support feature exploration within the model-predicted variation. We use a 300-million parameter multilingual version \cite{babu2021xls} of Wav2Vec2 \cite{baevski2020wav2vec}. For each synthesised utterance, we take the mean of all predicted Wav2Vec2 embeddings as a \textit{summary embedding} for the utterance. Feature exploration, which is discussed in Section \ref{sec:feature-exploration}, is carried out in a low-dimensional projection of these embeddings.

We additionally estimate the synthesised F0 contours using REAPER\footnote{https://github.com/google/REAPER} and extract functional features from the \texttt{GeMaps-v01b} feature set using OpenSmile \cite{opensmile}. We use these extracted features to evaluate whether the control features we identify correlate with known and interpretable acoustic features.

\subsection{Finding Controllable Features} \label{sec:feature-exploration}

We aim to discover features that, through follow-up fine-tuning of the model, can be controlled similarly to the other existing control features. At the core, the proposed method is based on the analysis of model variance --- in terms of the summary embeddings described in Section \ref{sec:representations} --- across a large number of synthesised samples (the \textit{analysis set}). But, the model can already be conditioned to generate acoustic variance in several ways: (1) through choice of target text, (2) choice of target speaker, (3) instructions included in the description prompt. So diversity in one or more of these inputs would naturally contribute to the observed model variance for the analysis set, thus affecting the results.

Initially, we explored several different compositions of the analysis set to evaluate what types of sets can support the proposed feature exploration. We reflect on the different results yielded by different compositions in Sections \ref{sec:t3}-\ref{sec:t3}. In total, we generate over 100,000 utterances across all analysis sets we evaluate. However, we found that the summary embeddings are inherently speaker- and text-dependent. This is true, no matter which layer of Wav2Vec2 we use to generate the embeddings (see Figure \ref{fig:diverse_set_everyone_5_pca_text_type}). For this reason, we focus our efforts only on fixed inputs: samples are generated using the same target text, target speaker and description prompt. \cite{lin2023utility} suggest that early layers of models like Wav2Vec2 contribute more to prosodic information structures than later layers. For Wav2Vec2 in particular, \textit{contribution} in the five tasks they evaluate comes mainly from the first four layers. We choose the 4th layer for the rest of our experiments based on our initial analysis.

After synthesising an analysis set, we project all synthesised summary embeddings into three dimensions using Principal Component Analysis (PCA). We perform this projection to make feature exploration tractable: the original embedding space is too complex to support the manual feature search we propose. Since we limit our analysis to fixed inputs, we assume that most of the observed variation arises from differences in prosody. We expect variance stemming from discrete variables (like speaking styles) to appear as distinct clusters in the PCA space, while continuous variables (like mean pitch) may manifest as gradients within those clusters. By manually inspecting the PCA projections, we determine whether the observed variation reflects discrete or continuous control features, which informs how we enrol the discovered features before further fine-tuning.

Given distinct clustering in PCA space, corresponding to a discrete control variable, we apply K-means to compute mean Wav2Vec2 embeddings for each cluster. Each training utterance is then labelled according to the lowest cosine distance to each cluster mean. In cases where the PCA reveals a continuous gradient, we discretise the synthesised samples into $n$ bins along the principal axis of variation. As with discrete variables, we compute mean embeddings for these bins and assign training labels based on cosine distance. The flexibility of prompt-based control allows us to express these cluster or bin labels as textual prompts that reflect their observed acoustic characteristics.

We adopt an iterative feature enrolment strategy: introducing only one new control label at a time before fine-tuning the model again. After each fine-tuning step, we repeat the analysis to capture the remaining variation not yet explained by previously enrolled control features.

\section{Results}
\subsection{Initial Hyperparameter tuning}\label{sec:hyperparams}

\begin{figure}[ht]
    \centering
    \includegraphics[width=\linewidth]{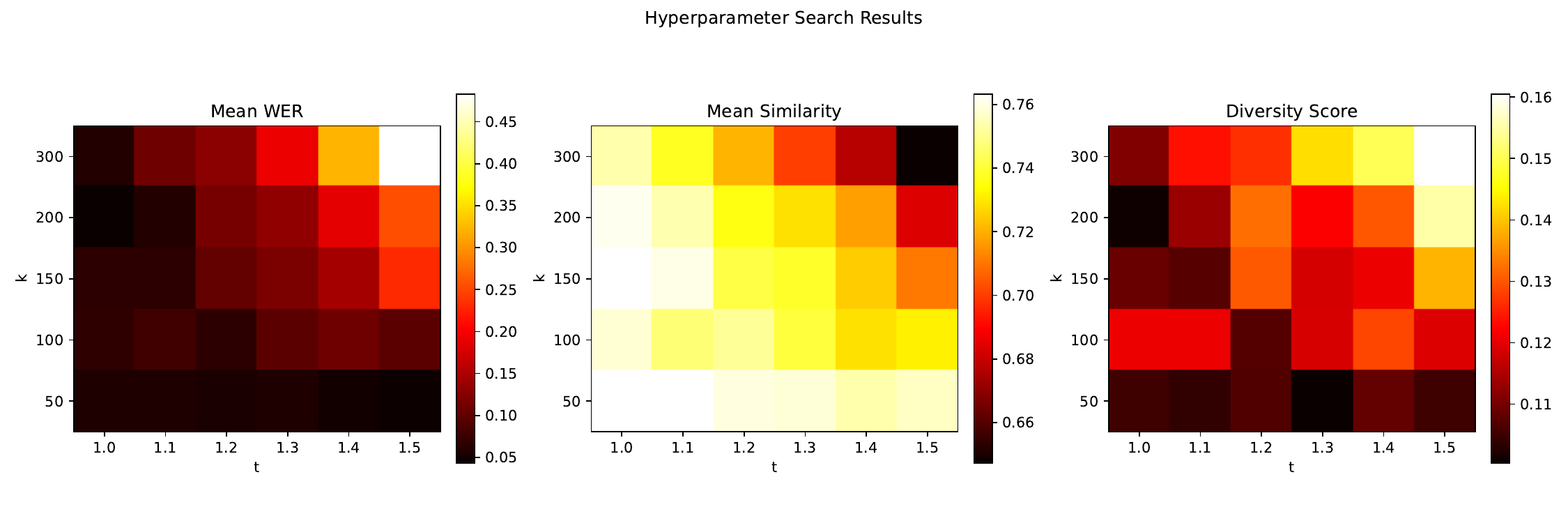}
    \caption{An overview of how the choice of temperature and $k$ affects the output distribution}
    \label{fig:hparam_heatmap}
\end{figure}

Several hyperparameters contribute to the plausible range of the model variation, and during inference, we can control how the model samples from the distribution over possible outputs. Primarily, we can do this by configuring the temperature ($\tau$) and through the choice of $k$ for $k$-sampling. Increasing temperature leads to less likely continuations being sampled more frequently, thus leading to higher variation in the output. Increasing $k$ forces the model to consider more candidates for continuations. We demonstrate what effect the choice of these hyperparameters has on the output in Figure \ref{fig:hparam_heatmap}. The third sub-figure shows the \textit{diversity score} of the set of generated Wav2Vec2 embedding summaries for different values of $\tau$ and $k$. A \textit{higher} diversity score corresponds to lower pairwise cosine similarity, higher cosine distance conversely, across the set:
\begin{align*}
    \text{diversity score} = \frac{1}{N(N-1)}\sum_{i=1}^{N}\sum_{j=1}^{N} (1 - \cos(\mathbf{e}_i, \mathbf{e}_j)), \quad i \neq j
\end{align*}
\noindent where $\mathbf{e}_i$ and $\mathbf{e}_j$ are the summary embeddings corresponding to the $i^{th}$ and $j^{th}$ utterance, respectively and $N$ is the total number of utterances in the set. As Figure \ref{fig:hparam_heatmap} shows, the diversity score increases steadily --- suggesting higher summary embedding variance --- with higher values of $\tau$ and $k$. Higher diversity is, plausibly, conducive to discovering more varied control features. However, as Figure \ref{fig:hparam_heatmap} shows, there is a trade-off between diversity and how comprehensive the output is. First, we evaluate this in terms of how high the Word Error Rate (WER) is, as measured by using a Wav2Vec2-based ASR system fine-tuned on Icelandic\footnote{https://huggingface.co/language-and-voice-lab/wav2vec2-large-xlsr-53-icelandic-ep30-967h}. We then evaluate how similar the output speaker is to the ground truth speaker. We do this by measuring the cosine ditsance of all generated samples to a mean of real speaker embeddings for the corresponding speaker. We use Resemblyzer \cite{resemblyzer} to generate all speaker embeddings, and we use 300 randomly sampled utterances of each speaker to create these mean speaker embeddings. As figure \ref{fig:hparam_heatmap} shows, speaker similarity drops substantially as temperature increases. Variation derived from confused speaker identity can affect our analysis, so based on these results, we choose $\tau=1.2$ and $k=100$ for our experiments; otherwise, we do not change the values of any other generation hyperparameters.

\subsection{Baseline Models} \label{sec:base-models}
\begin{table*}[ht]
    \centering
    \caption{Comparison of real and synthetic speakers (\textbf{T3-emotion}) in an objective evaluation of speaker consistency, intelligibility and diversity.} \label{tab:eval}
    \begin{tabular}{lllllll}
    \toprule
    \multicolumn{1}{c}{\multirow{2}{*}{\textbf{Speaker}}} & \multicolumn{2}{c}{\textbf{Speaker similarity}}                       & \multicolumn{2}{c}{\textbf{WER}}                              & \multicolumn{2}{c}{\textbf{Diversity score}}                        \\
    \multicolumn{1}{c}{}                         & \multicolumn{1}{c}{real} & \multicolumn{1}{c}{synth} & \multicolumn{1}{c}{real} & \multicolumn{1}{c}{synth} & \multicolumn{1}{c}{real} & \multicolumn{1}{c}{synth} \\
    \midrule
    Astrid & $0.78 \pm 0.04$ & $0.76 \pm 0.05$ & $0.02 \pm 0.05$ & $0.13 \pm 0.17$ & $0.13 \pm 0.06 $ & $0.20 \pm 0.06 $ \\
    Anders & $0.71 \pm 0.04$ & $0.71 \pm 0.05$ & $0.04 \pm 0.07$ & $0.12 \pm 0.15$ & $0.15 \pm 0.03 $ & $0.20 \pm 0.08 $ \\
    Ingrid & $0.77 \pm 0.05$ & $0.77 \pm 0.05$ & $0.04 \pm 0.09$ & $0.13 \pm 0.19$ & $0.14 \pm 0.05 $ & $0.19 \pm 0.06 $ \\
    Frida  & $0.72 \pm 0.04$ & $0.73 \pm 0.05$ & $0.05 \pm 0.10$ & $0.15 \pm 0.18$ & $0.14 \pm 0.03 $ & $0.21 \pm 0.11 $ \\
    Leif   & $0.75 \pm 0.06$ & $0.75 \pm 0.04$ & $0.03 \pm 0.07$ & $0.10 \pm 0.13$ & $0.12 \pm 0.07 $ & $0.20 \pm 0.08 $ \\
    Freya  & $0.75 \pm 0.04$ & $0.75 \pm 0.04$ & $0.04 \pm 0.11$ & $0.12 \pm 0.15$ & $0.14 \pm 0.03 $ & $0.21 \pm 0.09 $ \\
    Bjorn  & $0.71 \pm 0.04$ & $0.71 \pm 0.05$ & $0.06 \pm 0.10$ & $0.14 \pm 0.16$ & $0.12 \pm 0.02 $ & $0.23 \pm 0.10 $ \\
    \midrule
    All    & $0.74 \pm 0.04$ & $0.74 \pm 0.05$ & $0.04 \pm 0.07$ & $0.13 \pm 0.14$ & $0.13 \pm 0.05$ & $0.21 \pm 0.10$ \\
    \bottomrule    
    \end{tabular}
\end{table*}

Before initiating feature enrolment, we prepare two \textit{baseline} ParlerTTS models fine-tuned on Icelandic speech. We start with \textbf{T3-emotion}: our model that is exposed to emotion and emotional intensity labels during training. We fine-tune the model for a total of 36 epochs. We initially validate the quality of this model by comparing it to ground truth data in terms of speaker consistency, intelligibility and the overall diversity of the set. We generate 100 samples, using diverse description prompts, for each speaker and compare these to 100 randomly sampled real utterances per speaker in the training corpus. The results from this analysis are shown in Table \ref{tab:eval}. We use the same speech recognition model as we did in Section \ref{sec:hyperparams} to evaluate intelligibility and the same diversity and speaker similarity metric. In general, the synthetic speech is significantly ($\alpha < .05$) less intelligible according to a two-sample t-test ($t(700)=-15.2, p<.001$). We do observe that, as is typical for large speech language models, the model occasionally hallucinates a repetition in the middle of a word. We hypothesise that this is the main contributor to this decreased intelligibility. The synthetic data is also significantly more diverse than the real data $(t(700)=-18.9, p<.001)$. There is a possibility that model-induced hallucinations also contribute to this result. However, there is no statistical difference between the two sets in terms of speaker similarity $(t(700)=0.0, p=1.0)$. Based on initial inspection, we determined that a speaker similarity of $0.75$ or higher corresponded to perceptually high similarity to one of the real speakers. Several of the synthetic speakers are above this threshold, but we chose \textit{Ingrid} as our main speaker in our experiments.
\begin{figure}[t]
    \centering
    \includegraphics[width=0.9\linewidth]{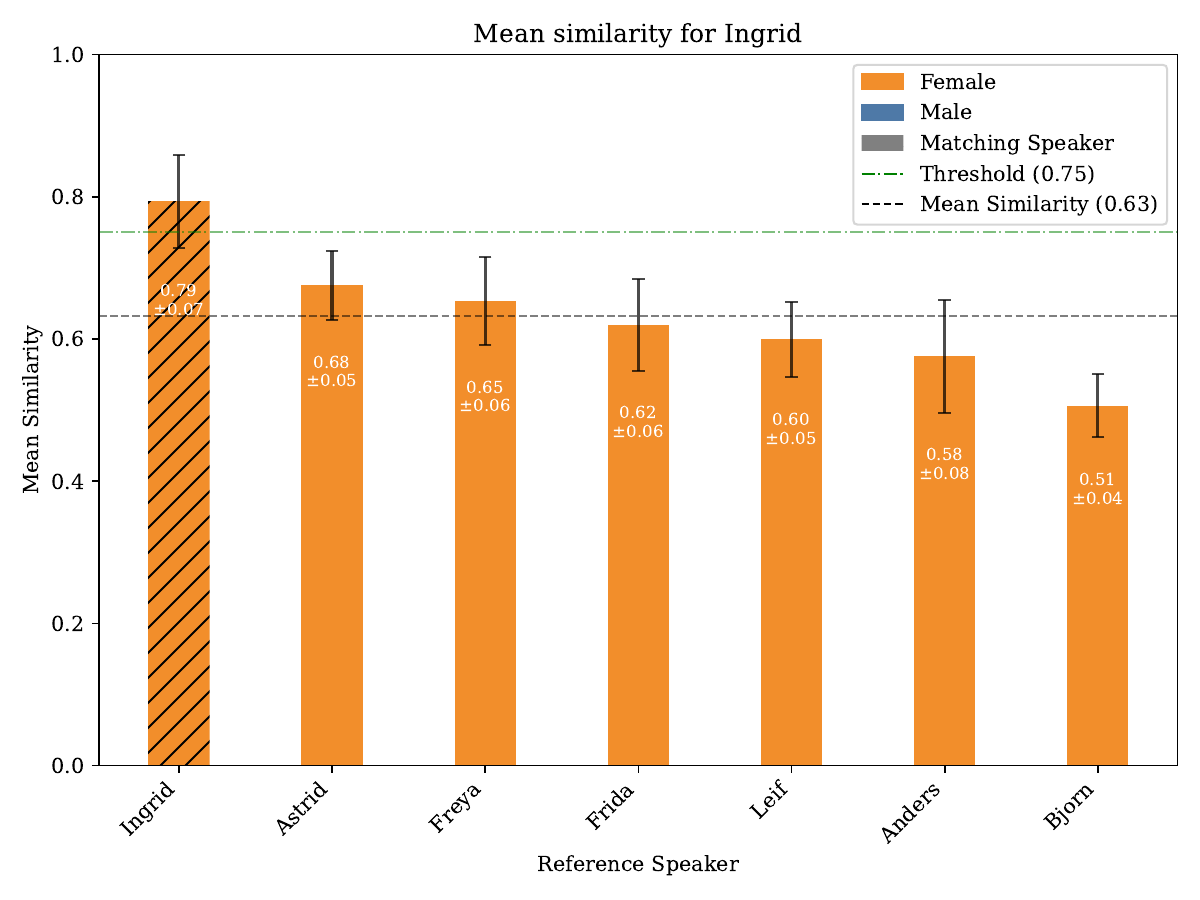}
    \caption{Speaker similarity of synthesised Ingrid to all real speakers}
    \label{fig:ingrid_speaker_sim_with_others}
\end{figure}

\begin{figure}[t]
    \centering
    \includegraphics[width=0.9\linewidth]{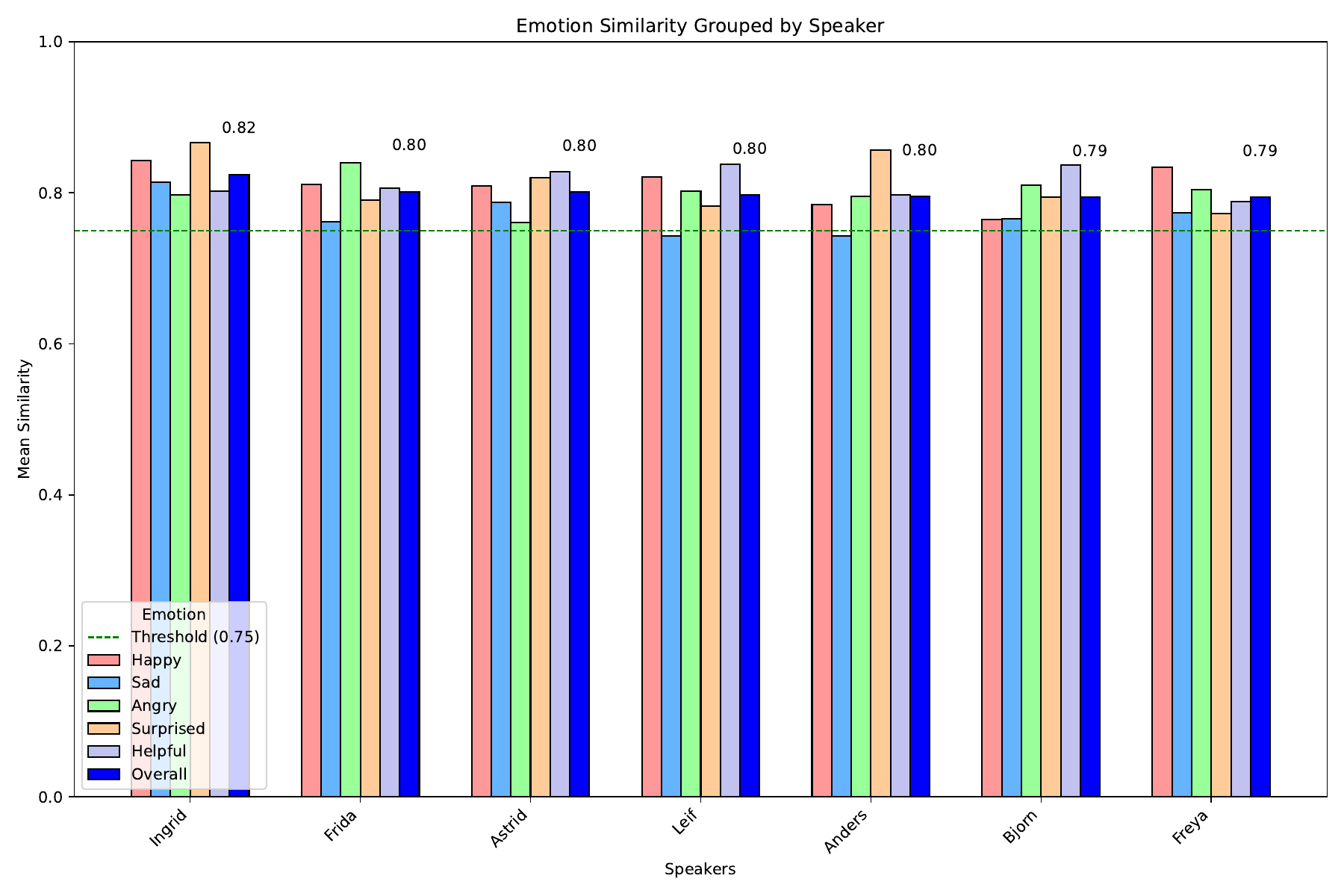}
    \caption{Grouped Emotion Similarity Chart}
    \label{fig:grouped_emotion_similarity}
\end{figure}

We perform a secondary \textit{neutral-speaking} trial to evaluate whether the speakers were consistently dissimilar from each other. We generate 1000 samples per speaker and determine, for each sample, which real speaker it most resembles. In all cases, the synthesised voices sounded most like themselves. Figure \ref{fig:ingrid_speaker_sim_with_others} shows the similarity of synthetic \textit{Ingrid} to all real speakers in the corpus. We further study the speaker-consistency of the model under different emotive classes. Here we measure the cosine similarity of speaker embeddings generated by the target speaker rendering a particular emotion class, with embeddings of model-generated utterances for the same speaker-emotion condition. Under these conditions, the model was consistently more similar to the target speaker than for neutral utterances. The mean cosine similarity for all speakers was above the previously determined $0.75$ threshold, as shown in Figure \ref{fig:grouped_emotion_similarity}. This, likely, suggests that the corpus speakers employ a specific set of prosodic features for rendering the different emotional categories, and that the model has learnt this mapping as well.

Our second model, \textbf{T3} is fine-tuned in the same way but is not exposed to emotional labels or intensities. We train \textbf{T3} for the same number of epochs as \textbf{T3-emotion}. We first perform feature enrolment using \textbf{T3}, as a hypothetically easier test case for the proposed method.

\subsection{T3 feature enrolment} \label{sec:t3}

\textbf{T3} is not exposed to any information about the emotional content of the utterances. Therefore, we hypothesise that the feature enrolment will yield controllable features relevant to the emotional content of the utterances. We first create a single-text, single-speaker analysis set, using \textit{Ingrid} as our target speaker, using the same description prompt (\textit{Ingrid sounds very clear and close to the microphone}). We synthesise 1,000 such samples, allowing the model to sample whichever prosodic and emotional rendition it deems probable. The generated set features a broad spectrum of prosody, reflecting on both emotional class and intensity. PCA reveals that The first two principal components account for $42.7\%$ of the total variance ($0.513$). We evenly sample and listen to 30 generated utterances across the first principal component axis. We determine the first principal correlated with the emotional intensity of the rendition, from low to high intensity. We discretise the range into three bins and create appropriate labels for each (1: \textit{low intensity}, 2: \textit{medium intensity}, 3: \textit{high intensity}). 

We create mean bin embeddings by taking the mean of 50 random samples from each bin, to re-label the training corpus according to lowest cosine distance. For example, we append \textit{Ingrid speaks at a very high intensity} to description prompts for utterances closest to the 3rd mean embedding. Table \ref{tab:percentage} lists which ground truth emotional types are paired with which cluster. We categorise any ground truth utterance with an emotional label higher than 3 as: \textit{Emotion - high intensity} and otherwise as: \textit{Emotion - low intensity}.
The table shows a substantial overlap of actual emotions across the three clusters, indicating only partial success in classifying emotional intensity.

We then fine-tune \textbf{T3} again, now using the instruction prompts modified to include the newly uncovered control labels. We limit further fine-tuning to \textit{Ingrid}, carrying out 40 epochs on the corpus subset. After fine-tuning, we generate three analysis sets, corresponding to the three new control labels, comprising 1,000 samples each. These three sets are analysed separately. After enrolling the new features, we find that the first two principal components account for, on average, $23.7\%$ of the total variance ($0.413$). This suggests that the method has reduced overall variance in the output.

\begin{table}[]
\centering
\caption{Clusters assigned to real utterances after second fine-tuning, categorised by ground-truth emotional labels } \label{tab:percentage}
\begin{tabular}{llll}
\toprule
\textbf{Assigned label}   & \multicolumn{1}{c}{\textbf{1}} & \multicolumn{1}{c}{\textbf{2}} & \multicolumn{1}{c}{\textbf{3}} \\
\midrule
Neutral                  & 59.3\% & 31.1\% & 9.6\%  \\
Emotion - low intensity  & 24.4\% & 48.3\% & 27.3\% \\
Emotion - high intensity & 5.3\%  & 36.5\% & 58.2\% \\
\bottomrule
\end{tabular}
\end{table}

\begin{figure}[t]
    \centering
    \includegraphics[width=\linewidth]{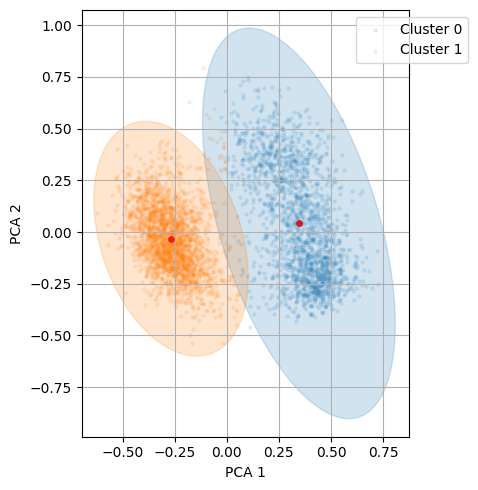}
    \caption{K-means clustering of projected embeddings from the \textit{T3 - Ingrid} subset. Cluster 0 consists of emotional renditions, while cluster 1 contains neutral renditions.}
    \label{fig:t3_ingrid_clusters}
\end{figure}

Interestingly, analyses of these sets repeatedly yield two clear clusters in PCA space as Figure \ref{fig:t3_ingrid_clusters} illustrates for one of the subsets. We perform 2-component K-means on the synthesised embeddings and observe that one cluster corresponds to the \textit{Neutral} corpus label while emotive speech is represented by the other, ranging from low-intensity to high as before. This is consistent across the three analysis sets. We create a mean embedding representing the neutral cluster (labelled $1$ in Table \ref{tab:percentage_2}) and two embeddings for high and low intensity utterances as before (labels $2$ and $3$ respectively). We then re-label the \textit{Ingrid} subset again according to cosine distance to cluster means. Separation based on these clusters results in a further reduction in mean total variance, across the two cluster labels, to $0.353$. Again we evaluate which ground-truth emotional intensities correspond to each cluster mean. The results of this are shown in Table \ref{tab:percentage_2}. We see a substantial reduction in the confusion of neutral and non-neutral utterances, validating the clustering approach. Although further fine-tuning stages may yield other control features, we stop the analysis of \textbf{T3} here.

\begin{table}[h]
\centering
\caption{Clusters assigned to real utterances after third fine-tuning, categorised by ground-truth emotional labels } \label{tab:percentage_2}
\begin{tabular}{llll}
\toprule
\textbf{Assigned label}   & \multicolumn{1}{c}{\textbf{1}} & \multicolumn{1}{c}{\textbf{2}} & \multicolumn{1}{c}{\textbf{3}} \\
\midrule
Neutral                  & 89.3\%                & 5.2\%                 & 5.5\%                 \\
Emotion - low intensity  & 12.3\%                & 49.1\%                & 38.6\%                \\
Emotion - high intensity & 1.4\%                 & 33.3\%                & 65.3\%    \\
\bottomrule
\end{tabular}
\end{table}

\subsection{\textbf{T3-emotion} feature enrolment} \label{sec:t3-emotion}
\begin{figure}[ht]
    \centering
    \includegraphics[width=\linewidth]{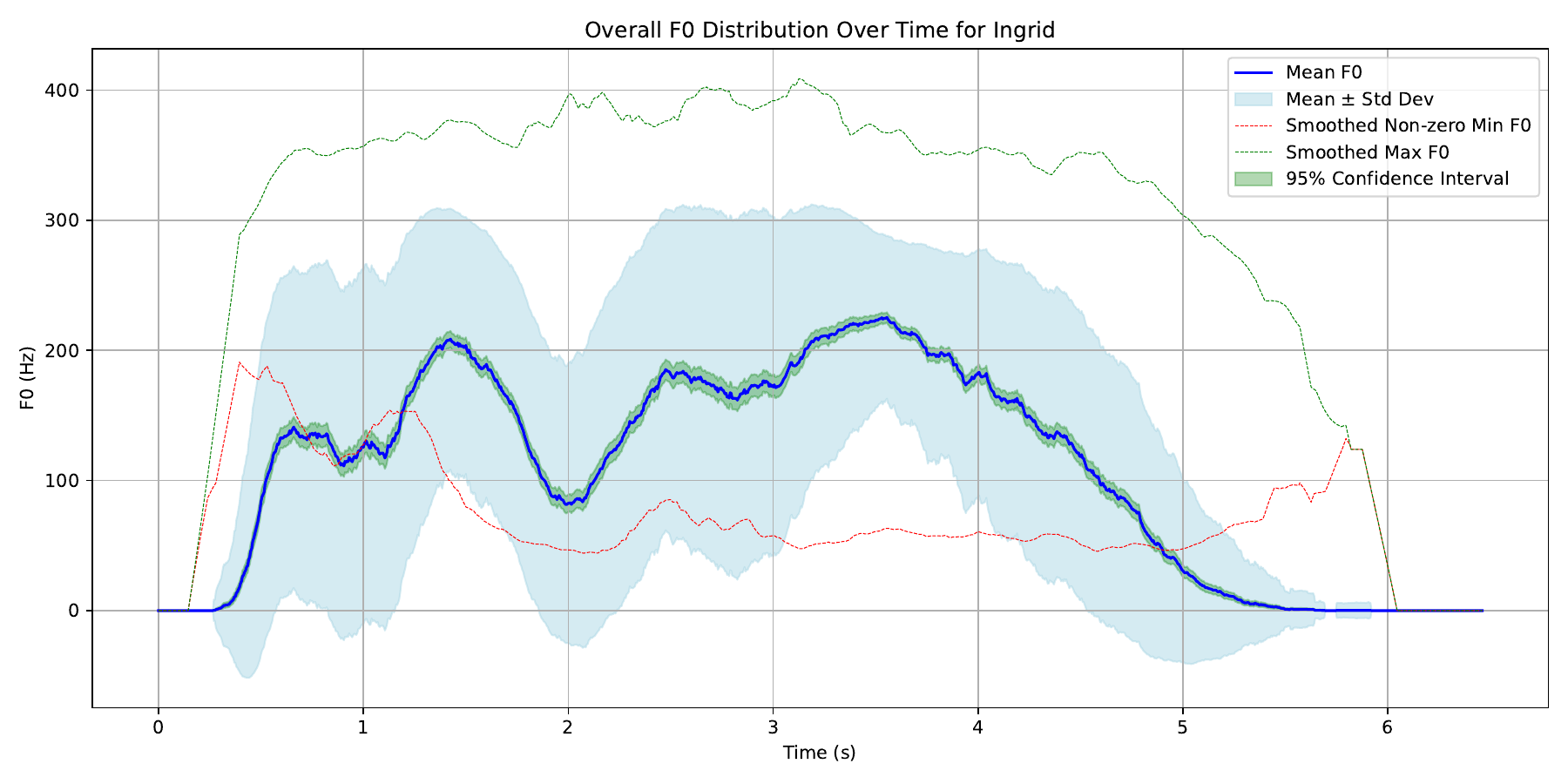}
    \caption{Overview of all generated \F0 contours for the single-speaker neutral emotion analysis set.}
    \label{fig:overall_f0_distribution_ingrid}
\end{figure}

The results from Section \ref{sec:t3} suggest that the proposed method can find and enroll unlabelled emotional features in the training corpus for secondary fine-tuning. We start our analysis \textbf{T3-emotion} of with a single-speaker, single-text neutral-emotion analysis set with the aim of finding prosodic features not directly related to emotional content. We generate 1000 samples in this case, using the same target text (\textit{Það var fallegt veður í gær og ég sá mörg dýr á leið minni í gegnum skóginn}\footnote{E: \textit{The weather yesterday was beautiful and I saw many animals on my way through the forest}}) and the same description prompt (\textit{Ingrid speaks in a neutral tone without any particular emotion. Ingrid's voice is clear, and she is close to the microphone.}). There are several plausible prosodic renditions for this text, but Figure \ref{fig:overall_f0_distribution_ingrid} demonstrates that the model is highly consistent in its rendering. The figure shows a summary of the 1,000 different \F0 contours for the generated samples. As this overview indicates, a large majority of those contours follows the same exact pattern.  

\begin{figure}[ht]
    \centering
    \includegraphics[width=0.7\linewidth]{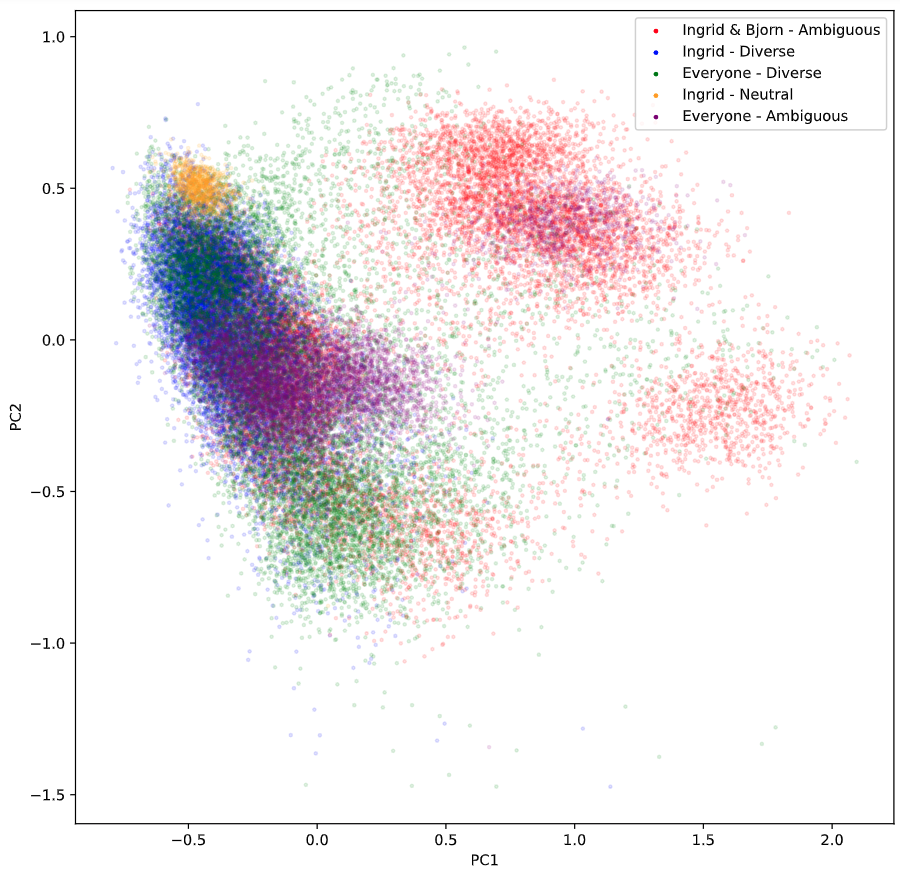}
    \caption{PCA projection of different \textbf{T3-emotion} renditions for the same target text, colored by analysis set.}
    \label{fig:all_emotional_clusters}
    \vspace{-0.5cm}
\end{figure}

We hypothesise that the model has learned a specific mapping for the neutral emotional label and will, therefore, not deviate too far from what it considers to be probable for that input description. When we compare this set to other renditions of the same text, from more diverse analysis sets, we see that the neutral single-speaker set is very limited in variation, as Figure \ref{fig:all_emotional_clusters} illustrates. This figure is a projection of all renditions of the same target text across different analysis sets. Projections of utterances from the initial neutral analysis set, \textit{Ingrid - Neutral}, are highly consistent with each other compared to other sets. 

\begin{figure}[ht]
    \centering
    \includegraphics[width=\linewidth]{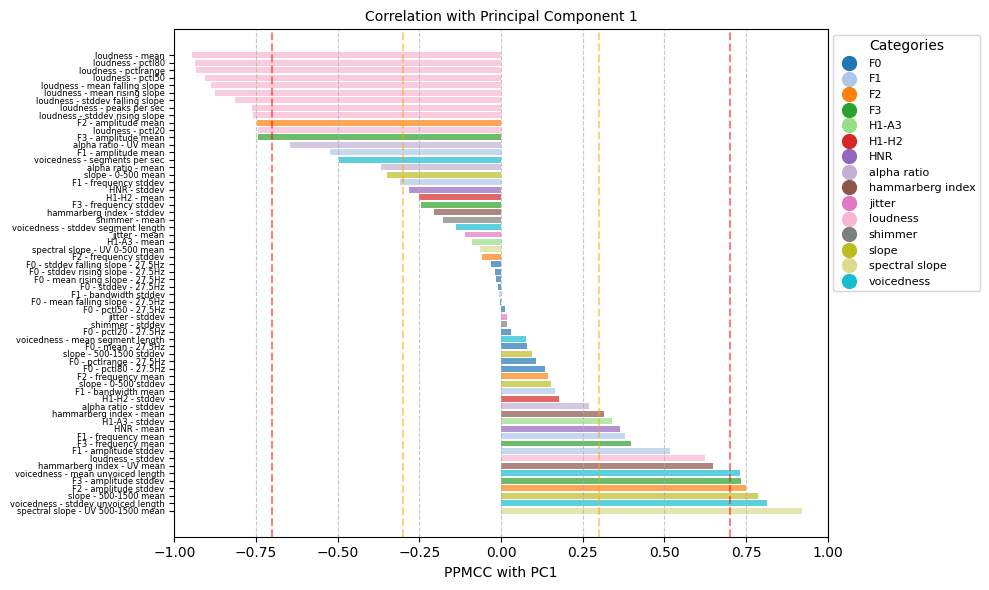}
    \caption{Correlation of \texttt{GeMaps-v01b} features with principal component of \textbf{diverse-ingrid}}
    \label{fig:feature_contribution_ingrid}
    \vspace{-0.5cm}
\end{figure}

We therefore decide to instead prompt the model without an emotional label. We hypothesise that omitting an emotion label would leave the model free to determine, from its underlying distribution, a single emotion label or a mixture of labels to render the text in. Again, we used \textit{Ingrid} as our target speaker. We did observe a higher level of variance for this analysis set compared to the neutral test case. We determine which of the \texttt{GeMaps-v01b} features best correlate with the first principal component. The results of this are shown in Figure \ref{fig:feature_contribution_ingrid}. There is a strong negative correlation between features related to perceived loudness and the first principal. Based on manual analysis, we determine that this correlation is explained by variation in recording condition across the Talromur-3 corpus. Since the first iteration of the feature enrolment method fails to yield a feature relevant to prosody we stop the analysis here.

\section{Discussion and Conclusion}
Our evaluation of \textbf{T3} shows that the proposed method can uncover unlabelled prosodic features in the training corpus by analysing thousands of samples generated for the same inputs. We also demonstrate that simple classification, based on PCA projection of utterance-level embeddings, can be used to enrol these features to the model through secondary fine-tuning. However, this approach does not generalise to \textbf{T3-emotion}, where the method instead reveals a correlation reflecting on the recording environment in which the corpus was originally created. This highlights the method's sensitivity to any variation in the speech signal, which can hinder the discovery of meaningful prosodic features. In our study, we employ mean embeddings generated by a multilingual Wav2Vec2 for utterance represenations. Alternative representations, which were not explored in the current work, may be less affected by non-verbal variation and, therefore, more suitable for the proposed approach.

\section{Acknowledgements}
This work was supported in part by Huawei and the UKRI Centre for Doctoral Training in Natural Language Processing, funded by the UKRI (grant EP/S022481/1) and the University of Edinburgh, School of Informatics and School of Philosophy, Psychology \& Language Sciences.

\bibliographystyle{IEEEtran}
\bibliography{mybib}

\end{document}